\documentclass[runningheads]{llncs}
\pdfoutput=1
\usepackage{times}
\usepackage{helvet}
\usepackage{courier}
\usepackage{xspace}
\usepackage{latexsym}
\usepackage{verbatim}
\usepackage{amsmath}

\usepackage{amsthm}
\usepackage{amssymb}
\usepackage{color}
\usepackage{enumerate}
\usepackage{tikz}
\usetikzlibrary{shapes,arrows}

\usepackage{booktabs}
\usepackage{alltt}
\usepackage{caption}

\usepackage{graphicx}
\graphicspath{{figures/}}
\usepackage{subfig}

\usepackage{url}
\def\mp{\mathcal{D}}
\def\mr{\mathcal{R}}
\def\mf{\mathcal{F}}
\def\mg{\mathcal{G}}

\newcommand{\rarc}{\textit{All Rules Cancelled}\xspace}
\newcommand{\rbt}{\textit{Backchain True}\xspace}
\def\vph{\ensuremath{\varphi}}
 \def\cS{\ensuremath{\mathcal{A}}}
 
 \def\cT{\ensuremath{\mathcal{T}}}
 \def\cP{\ensuremath{\mathcal{P}}}
 \def\cL{\ensuremath{\mathcal{L}}}
 
 \def\cT{\ensuremath{\mathcal{T}}}

\def\at{\mathcal{A}}
\def\nt{\mathit{not\;}}

\def\ar{\leftarrow}

\def\lrar{\longrightarrow}
 \def\beq{\begin{equation}}
 \def\eeq#1{\label{#1}\end{equation}}
 \def\ba{\begin{array}}
 \def\ea{\end{array}}

\def\smodels{{\sc smodels}}
\def\dlvhex{{\sc dlvhex}}

 \newcommand{\dpll}{\textsc{dpll}}
\newcommand{\ol}{\overline}

\newcommand{\rup}{\ensuremath{\text{\textit{UnitPropagate}}}}
\newcommand{\rp}{\ensuremath{\text{\textit{Propagate}}}}

\newcommand{\rupl}{\textit{Unit Propagate Learn}}
\newcommand{\rl}{\textit{Learn}}
\newcommand{\rll}{\textit{Learn Local}}
\newcommand{\rlg}{\textit{Learn Global}}
\newcommand{\dec}{{\Delta}}
\newcommand{\rd}{\ensuremath{\text{\textit{Decide}}}}
\newcommand{\rf}{\ensuremath{\text{\textit{Fail}}}}
\newcommand{\rfarc}{\ensuremath{\text{\textit{Fail ARC}}}}
\newcommand{\rfbt}{\ensuremath{\text{\textit{Fail BT}}}}
\newcommand{\rb}{\ensuremath{\text{\textit{Backtrack}}}}
\newcommand{\rbarc}{\ensuremath{\text{\textit{Backtrack ARC}}}}
\newcommand{\rbbt}{\ensuremath{\text{\textit{Backtrack BT}}}}
 \newcommand{\runf}{\textit{Unfounded}}
 \newcommand{\runfm}{\textit{Unfounded Modular}}
\newcommand{\fail}{\hbox{$\bot$}}

\newcommand{\as}{\ensuremath{\text{\textit{\sc{as}}}}}

\newcommand{\sm}{\ensuremath{\text{\textit{\sc{sm}}}}}
\newcommand{\am}{\ensuremath{\text{\textit{\sc{am}}}}}
\newcommand{\ams}{\ensuremath{\text{\textit{\sc{ams}}}}}

\newcommand{\amsl}{\ensuremath{\text{\textit{\sc{amsl}}}}}

\newcommand{\smmod}{\textit{\sc{msm}}}
\newcommand{\smmodl}{\textit{\sc{msml}}}

\newcommand{\dps}{\ensuremath{\text{\textit{\sc{dp}}}}}

\newcommand{\bds}{\mathit{Bodies}}

\def\prop#1#2{

\smallskip
\noindent
{\bf Proposition~\ref{#1}} {\it{#2}}}

\def\thm#1#2{

\smallskip
\noindent
{\bf Theorem~\ref{#1}} {\it{#2}}}

\title{Abstract Modular Systems and Solvers}

\begin{document}
\author{Yuliya Lierler\inst{1} \and
Miroslaw Truszczynski\inst{2}
}
\titlerunning{Abstract Modular Systems and Solvers}
\authorrunning{Y.~Lierler and M.~Truszczynski}
\setcounter{page}{143}

\institute{ University of Nebraska at Omaha\\
\email{ylierler@unomaha.edu}
\and
University of Kentucky\\
\email{mirek@cs.uky.edu}
}
\maketitle
\begin{abstract}
Integrating diverse formalisms into modular knowledge representation
systems offers increased expressivity, modeling convenience and 
computational benefits. 
We introduce concepts of 
\emph{abstract modules} and \emph{abstract modular systems}
to study general principles behind the design 
and analysis of model-finding programs, or \emph{solvers}, for
integrated heterogeneous multi-logic systems. 
We show how abstract modules and 
abstract modular systems give rise to \emph{transition systems}, 
which are a natural and convenient representation of solvers pioneered
by the SAT community. We 
illustrate our approach by showing how it applies to answer set 
programming and propositional logic, and to multi-logic systems based on 
these two formalisms.
\end{abstract}

\section{Introduction}
Knowledge representation
and reasoning (KR) is concerned with developing formal languages 
and logics to model knowledge, and with designing and implementing 
corresponding automated reasoning tools. The choice of specific 
logics and tools depends on the type of knowledge to be represented 
and reasoned about. Different logics are suitable for common-sense 
reasoning, reasoning under incomplete information and uncertainty, 
for temporal and spatial reasoning, and for modeling and solving 
boolean constraints, or constraints over 
larger, even continuous domains. In applications in areas such as 
distributed databases, semantic web, hybrid constraint modeling and 
solving, to name just a few, several of these aspects come to play. 
Accordingly, often diverse logics have to be accommodated together. 
Similar issues arise in research on multi-context systems where the
major task is to
model \emph{contextual} information and the flow of information 
among \emph{contexts}~\cite{mcc87,Giunchiglia93}. The contexts
are commonly modeled by theories in some logics.

Modeling convenience is not the only reason why diverse logics are 
combined into modular hybrid KR systems. 
Another major motivation is to exploit in reasoning the transparent 
structure that comes from modularity, computational strengths of 
individual logics, and synergies that may arise when they are put 
together. Constraint logic programming \cite{jm94} and 
satisfiability modulo theories (SMT) \cite{nie06,BarretSST08} 
are well-known examples of formalisms stemming directly from such 
considerations. More recent examples include constraint answer set 
programming (CASP) \cite{lier12aaai}, which integrates answer set 
programming (ASP) \cite{gel88,mar99,nie99}) with constraint modeling 
languages \cite{ros08}, 
and
``multi-logic'' formalisms PC(ID)~\cite{mar08}, SM(ASP) \cite{lt2011} 
and ASP-FO \cite{dltj12} that combine modules expressed as
logic theories under the classical
semantics with 
modules given as answer-set programs.
 
The key computational task arising in KR is that of model generation. 
Model-generating programs or \emph{solvers}, developed 
in satisfiability (SAT) and ASP proved to be effective in a 
broad range of KR applications. Accordingly, model generation is of 
critical importance in modular multi-logic systems. Research on 
formalisms listed above resulted in fast solvers that demonstrate
gains one can obtain from their heterogeneous nature. However, the
diversity of logics considered and low-level technical details of 
their syntax and semantics obscure general principles that are
important in the design and analysis of solvers for multi-logic 
systems.

In this paper we address this problem by proposing a language
for talking about modular multi-logic systems that (i) abstracts away the 
syntactic details, (ii) is expressive enough to capture various concepts 
of inference, and (iii) is based only on the weakest assumptions concerning 
the semantics. The basic elements of this language are 
\emph{abstract modules}.  Collections of abstract modules constitute
\emph{abstract modular systems}. We define the semantics of
abstract modules and show that they 
provide a uniform language capable of capturing different 
logics, diverse inference mechanisms, and their modular
combinations. Importantly, abstract modules and abstract modular systems 
give rise to \emph{transition systems} of the type introduced by 
Nieuwenhuis, Oliveras, and Tinelli~\cite{nie06} in their study of SAT and SMT 
solvers. We show that as in that earlier work, our transition systems 
provide a natural and convenient representation of solvers for abstract 
modules and abstract modular systems. 
We demonstrate that they lend themselves
well to extensions that capture such important solver design techniques 
as learning (which here comes in two flavors: \emph{local} that is 
limited to single modules, and \emph{global} that is applied across 
modules). 
Throughout the paper, we illustrate our approach by showing
how it applies to propositional logic and answer set programming, and
to multi-logic systems based on these two formalisms.  

The results of our paper show that abstract modular systems and the
corresponding abstract framework for describing and analyzing algorithms 
for modular declarative programming tools relying on multi-logics are 
useful and effective conceptualizations that can contribute to (i) 
clarifying computational principles of such systems and to (ii) the 
development of new ones.

The paper is organized as follows.
We start by introducing one of the main concepts in
the paper -- abstract modules. We then proceed to formulating an
algorithm (a family of algorithms) for finding models of such modules. We
use an abstract transition system stemming from
the framework by Nieuwenhuis et al.~\cite{nie06} for this purpose.
Section~\ref{sec:ams} presents the definition of an abstract modular
system and a corresponding solver based on backtrack search. We
then discuss how this solver maybe augmented by such advanced SAT
solving technique as learning. Section~\ref{sec:related} provides an
account on related work.

\section{Abstract Modules}

Let $\sigma$ be a fixed finite vocabulary (a set of propositional 
atoms). A \emph{module} over the
vocabulary $\sigma$ is a directed graph~$S$ whose nodes are $\bot$ and
all consistent sets of literals, and each edge is of the form $(M,\bot)$
or $(M,Ml)$, where $l\notin M$ and $Ml$ is a shorthand for $M\cup \{l\}$. 
If $S$ is a module, we write $\sigma(S)$ for its vocabulary.
 For a set $X$ of literals, we denote $X^{+}=\{a\colon a
\in X\}$ and $X^{-}=\{a\colon \neg a\in X\}$. 

Intuitively, an edge  $(M,Ml)$ in a module indicates that the module 
supports inferring $l$ whenever all literals in $M$ are given. An edge 
$(M,\bot)$, $M\not=\emptyset$, indicates that there is a literal $l\in M$ 
such that a derivation of its dual $\bar{l}$ (and hence, a derivation of a 
contradiction) is supported by the module, assuming the literals in $M$ 
are given. 
Finally, the edge $(\emptyset,\bot)$ indicates that the module is 
``explicitly'' contradictory. 

A node in a module is {\em terminal} if no edge leaves it. A terminal 
node that is consistent and complete is a \emph{model node} of the
module. 
A set $X$ of atoms is a \emph{model} of a module $S$ if for some model node $Y$
in $S$, $X\cap\sigma(S)=Y^{+}$. Thus, models of modules are not restricted to
the signature of the module. Clearly, for every model node $Y$ in $S$, 
$Y^+$ is a model of $S$.  


\tikzstyle{node}=[text centered]
\tikzstyle{vertex}=[draw, rectangle,  text centered,
  rounded corners]
\tikzstyle{line}=[draw, -latex']

\begin{figure}[ht]
\centering{
{\footnotesize
\begin{tikzpicture}

\node [vertex] (n4) at (2,1.5) {$a$};
\node [node] (n10) at (2,2.4) {(a)};

\node [vertex] (n1) at (2.8,1) {$\emptyset$};
\node [vertex] (n11) at (2.8,2) {$\bot$};

\node [vertex] (n5) at (3.6,1.5) {$\neg a$};

\node [vertex] (n6) at (4.6,1.5) {$a$};
\node [node] (n12) at (4.6,2.4) {(b)};

\node [vertex] (n13) at (5.4,2) {$\bot$};
\node [vertex] (n2) at (5.4,1) {$\emptyset$};

\node [vertex] (n7) at (6.2,1.5) {$\neg a$};

\node [vertex] (n8) at (7.2,1.5) {$a$};
\node [node] (n14) at (7.2,2.4) {(c)};

\node [vertex] (n15) at (8,2) {$\bot$};
\node [vertex] (n3) at (8,1) {$\emptyset$};

\node [vertex] (n9) at (8.8,1.5) {$\neg a$};

\path [line] (n1) -- (n4);
\path [line] (n5) -- (n11);
\path [line] (n7) -- (n13);
\path [line] (n3) -- (n9);
\path [line] (n9) -- (n15);
\end{tikzpicture}
}
}
\caption{Three modules over the vocabulary $\{a\}$.}
\label{fig:modex1}
\end{figure}
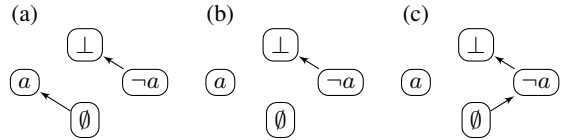

A module $S$ \emph{entails} a formula $\vph$, written $S\models\vph$,
if for every model $I$ of $S$ we have $I\models\vph$. It is immaterial
what logic the formula $\vph$ comes from as long as 
(i) the vocabulary of the logic is a subset of the vocabulary of $S$, and
(ii) the semantics of the logic is given by a satisfiability relation
\hbox{$I\models\vph$}.
A module $S$ \emph{entails} a formula $\vph$ wrt a set $M$ of literals 
(over the same vocabulary as $S$), written $S\models_M \vph$, if for 
every model $I$ of $S$ such that $M^+\subseteq I$ and $M^-\cap I=\emptyset$, $I\models \vph$.

Clearly, if two modules over the same signature have the same model nodes,
they have the same models. Semantically the three modules in Figure 
\ref{fig:modex1} are the same. They have the same models (each has $\{a\}$ 
as its only model in the signature of the module) and so they entail the 
same formulas.
We call modules with the same models \emph{equivalent}.

Modules represent more than just the set of their models. As already
suggested above, the intended role of edges in a module is to represent allowed 
``local'' inferences. 
For instance, given 
the empty set of literals, the first module in Figure \ref{fig:modex1} 
supports inferring $a$ and the
third module $\neg a$. In the latter case, the inference is not ``sound'' 
as it contradicts the semantic information in the module as that module 
does not entail $\neg a$ with respect to the empty set of literals.  

Formally, an edge from a node $M$ to a node $M'$ in a module $S$ is 
\emph{sound} if $S\models_M M'$.\footnote{In the paper, we sometimes 
identify a set of literals with the conjunction of its elements. Here
$M'$ is to be understood as the conjunction of its elements.} 
Clearly, if $M'$ has the form $Ml$ then $S\models_M M'$ if and only
if $S\models_M l$. Similarly, if $M'=\bot$ then $S\models_M M'$ if and only
if no model of $S$ is consistent with $M$ (that is, contains $M^{+}$
and is disjoint with $M^{-}$).
A module is \emph{sound} if all of its 
edges are sound, that is, if all inferences supported by the 
module are sound with respect to the semantics of the module given by 
its set of models. 
The modules in Figures \ref{fig:modex1}(a) and (b)
are sound, the one in Figure \ref{fig:modex1}(c) is not. Namely, the 
inference of $\neg a$ from $\emptyset$ is not sound.

Given two modules $S$ and $S'$ over the same vocabulary, we say that
$S$ is \emph{equivalently contained} in $S'$, $S\sqsubseteq S'$, if 
$S$ and $S'$ are equivalent (have the same model nodes) and the set 
of edges of $S$ is a subset of the set of edges of $S'$.  Maximal (wrt
$\sqsubseteq$) sound 
modules are called \emph{saturated}.
We say that an edge from a node $M$ to $\bot$ in a module $S$ is 
\emph{critical} if $M$ is a complete and consistent set of literals
over $\sigma(S)$. 
 The following properties are 
evident.

\begin{proposition}
\label{prop:sat}
Every two modules over the same signature and with the same critical edges 
are equivalent. For a saturated module $S$, every sound module with the 
same critical edges as $S$ is equivalently contained in $S$. A module $S$
is saturated if and only if it is sound and for every set $M$ of literals 
and for every literal $l\not\in M$, $(M,Ml)$ is an edge of $S$ whenever $S\models_M l$.
\end{proposition}

Clearly, only the module in Figure \ref{fig:modex1}(a) is saturated. The
other two are not. The one in (b) is not maximal with respect to the 
containment relation, the one in (c) is not sound. We also note that
all three modules have the same critical edges. Thus, by Proposition
\ref{prop:sat}, they are equivalent, a property we already observed 
earlier. Finally, 
the module in Figure \ref{fig:modex1}(b)
is equivalently contained in the
module in Figure \ref{fig:modex1}(a). 


\begin{figure}[ht]
\centering{
{\footnotesize
\begin{tikzpicture}
\node [vertex] (n2) at (3.0,0.75) {$a$};
\node [vertex] (n6) at (3.0,1.5) {$a~ b$};

\node [vertex] (n7) at (4.0,1.5) {$a ~\neg b$};
\node [vertex] (n5) at (4.0,0.75) {$\neg b$};

\node [vertex] (n0) at (4.4,0) {$\emptyset$};
\node [vertex] (n17) at (4.4,2.25) {$\bot$};

\node [vertex] (n3) at (5.0,0.75) {$\neg a$};
\node [vertex] (n8) at (5.0,1.5) {$\neg a~ b$};

\node [vertex] (n4) at (6.2,0.75) {$b$};
\node [vertex] (n9) at (6.2,1.5) {$\neg a~ \neg b$};

\path [line] (n2) -- (n7);
\path [line] (n3) -- (n8);
\path [line] (n4) -- (n8);
\path [line] (n5) -- (n7);
\path [line] (n6) -- (n17);
\path [line] (n9) -- (n17);

\end{tikzpicture}
}
}
\caption{An abstract module over the vocabulary $\{a,b\}$ related
to the theory \eqref{e:ext1}.}
\label{fig:modex2}
\end{figure}
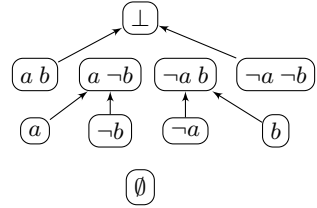

In practice, modules (graphs) are specified by means of theories and 
logics (more precisely, specific forms of inference in logics). For 
instance, a propositional theory $T$ over a vocabulary $\sigma$ and 
the inference method given by the 
classical concept of entailment determine a module over $\sigma$ in
which (i) $(M,Ml)$ is an edge  
if and only if $T\cup M \models l$; and (ii) $(M,\bot)$ is an edge
if and only if no model of $T$ is consistent with $M$. 
Figure~\ref{fig:modex1}(a)
shows the module  determined in this way 
by the theory consisting of the clause $a$.
Similarly, Figure~\ref{fig:modex2}
presents such a  module for the theory 
\beq
\ba c
a\lor b,\quad 
\neg a\lor \neg b.
\ea
\eeq{e:ext1} 
This module is saturated. 
Also, theory~\eqref{e:ext1} and the inference method given by the unit 
propagate rule, a classical propagator used in SAT 
solvers, determines this module. In other words, for the theory~\eqref{e:ext1} 
the unit propagation rule captures entailment.

We say that a 
module $S$ is {\em equivalent} to a theory~$T$ in 
some logic if the models of $S$ coincide with the models of~$T$.
Clearly, the module in Figure~\ref{fig:modex2} is equivalent to the
propositional
 theory~\eqref{e:ext1}.

Modules are not meant for modeling. 
Representations by means of logic theories are usually more
concise 
(the size of a module is exponential in the size of its vocabulary).
Furthermore, the logic languages 
align  closely with natural language, which facilitates modeling 
and makes the correspondence between logic theories and 
knowledge they represent  direct. Modules lack this connection to
natural language.

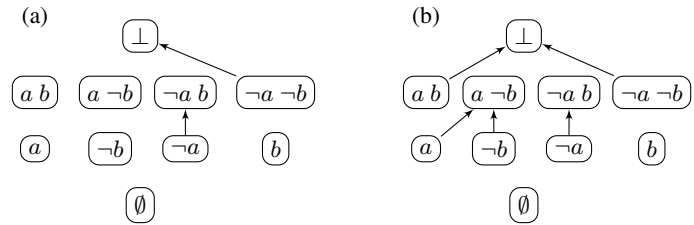
\begin{figure}[ht]
\centering
{\footnotesize
{\begin{tikzpicture}
\node [node] (lab1) at (3.0,2.5) {(a)};
\node [vertex] (n2) at (3.0,0.75) {$a$};
\node [vertex] (n6) at (3.0,1.5) {$a~ b$};

\node [vertex] (n7) at (4.0,1.5) {$a ~\neg b$};
\node [vertex] (n5) at (4.0,0.75) {$\neg b$};

\node [vertex] (n0) at (4.4,0) {$\emptyset$};
\node [vertex] (n17) at (4.4,2.25) {$\bot$};

\node [vertex] (n3) at (5.0,0.75) {$\neg a$};
\node [vertex] (n8) at (5.0,1.5) {$\neg a~ b$};

\node [vertex] (n4) at (6.2,0.75) {$b$};
\node [vertex] (n9) at (6.2,1.5) {$\neg a~ \neg b$};

\path [line] (n3) -- (n8);
\path [line] (n9) -- (n17);

\node [vertex] (n12) at (8.2,0.75) {$a$};
\node [vertex] (n16) at (8.2,1.5) {$a~ b$};
\node [node] (lab2) at  (8.2,2.5) {(b)};

\node [vertex] (n217) at  (9.1,1.5) {$a ~\neg b$};
\node [vertex] (n15) at  (9.1,0.75) {$\neg b$};

\node [vertex] (n10) at  (9.5,0) {$\emptyset$};
\node [vertex] (n117) at (9.5,2.25) {$\bot$};

\node [vertex] (n13) at (10.1,0.75) {$\neg a$};
\node [vertex] (n18) at (10.1,1.5) {$\neg a~ b$};

\node [vertex] (n14) at (11.2,0.75) {$b$};
\node [vertex] (n19) at (11.2,1.5) {$\neg a~ \neg b$};

\path [line] (n12) -- (n217);
\path [line] (n13) -- (n18);
\path [line] (n15) -- (n217);
\path [line] (n16) -- (n117);
\path [line] (n19) -- (n117);

\end{tikzpicture}
}
}
\caption{Two abstract modules over the vocabulary $\{a,b\}$
related to the logic program~(\ref{eq:pr}).}
\label{fig:modex3}
\end{figure}

The power of modules comes from the fact that they provide a uniform, 
syntax-independent way to describe theories and inference methods
stemming from 
\emph{different} logics. 
For instance, they represent equally well
both propositional theories 
and logic programs under the answer-set semantics. Indeed, let us consider
the logic program 
\beq
\ba{l}
\{a\},\\
b\ar \nt a,
\ea
\eeq{eq:pr}
where $\{a\}$ represents the so-called \emph{choice rule}
\cite{sns02}.
This program has two answer sets $\{a\}$ and $\{b\}$. Since these are
also the only two models of the propositional theory (\ref{e:ext1}),
it is clear that the module in Figure \ref{fig:modex2} represents 
the program~(\ref{eq:pr}) and the reasoning mechanism of entailment with
respect to its answer sets.
Two other 
modules associated with program~\eqref{eq:pr} are given in Figure 
\ref{fig:modex3}. The module in Figure~\ref{fig:modex3}(a) represents 
program (\ref{eq:pr}) and the reasoning on programs based 
on \emph{forward chaining}; we call this module~$M_{\mathit{fc}}$. We 
recall that given a set of literals, forward chaining supports the 
derivation of the head of a rule whose body is satisfied. We note that 
the module $M_{\mathit{fc}}$ is not equivalent to program (\ref{eq:pr}).
Indeed, $\{a, b\}$ is a model of $M_{\mathit{fc}}$ whereas it is not  an
answer set of (\ref{eq:pr}). This is due to the fact that the critical 
edge from $a\,b$ to $\bot$ is unsupported by forward chaining and is 
not present in $M_{\mathit{fc}}$. On the other hand, all edges due
to forward chaining are sound both in the module in Figure
\ref{fig:modex2}, which we call $M_e$, and $M_{\mathit{fc}}$.
In the next section we discuss a combination of inference rules that 
yields a reasoning mechanism subsuming forward chaining and resulting 
in a module, shown in Figure~\ref{fig:modex3}(b), that is equivalently 
contained in~$M_e$ and so, equivalent to the program~(\ref{eq:pr}). 
This discussion indicates that the language
of modules is flexible enough to represent not only the semantic 
mechanism of entailment, but also syntactically defined ``proof
systems'' --- reasoning mechanisms based on specific inference rules.

\section{Abstract Modular Solver: $\am_S$}

Finding models of logic theories and programs is a key computational 
task in declarative programming. 
Nieuwenhuis et al.~\cite{nie06}
proposed to use transition systems to describe search procedures 
involved in model-finding programs commonly called \emph{solvers}, and 
developed that approach for the case of SAT. Their transition system 
framework can express {\dpll}, the basic search procedure 
employed by SAT solvers, and its  enhancements such as 
conflict driven clause learning. 
Lierler~\cite{lier10} proposed a similar framework for specifying 
an answer set solver {\sc smodels}. Lierler and 
Truszczynski~\cite{lt2011} 
extended that framework to capture such modern ASP solvers as {\sc cmodels} 
and {\sc clasp}, as well as a PC(ID) solver {\sc minisat(id)}.

An abstract nature (independence from language and reasoning method selection)
 of modules introduced in this work and their
 relation to proof systems
makes them a convenient, broadly applicable tool to study and analyze 
solvers. In this section, we adapt the transition system framework of 
Nieuwenhuis et al.~\cite{nie06}
 to the case of abstract modules. We then illustrate
how it can be used
to define solvers for instantiations of abstract modules such as 
propositional theories under the classical semantics and logic programs 
under the answer-set semantics.

A {\em  state} relative to $\sigma$ is either a special state $\fail$
(fail state) or an \emph{ordered} consistent set~$M$ of literals over $\sigma$, some possibly
annotated by $\dec$, which marks them as {\em decision} literals.
 For instance, the
states relative to a singleton set~$\{ a\}$ of atoms are
$
\emptyset,\ \ a,\ \ \neg a, \ \ a^\dec,\ \neg a^\dec,\ \  
\fail.
$

Frequently, we consider a state $M$ as a set of literals,
ignoring both the annotations and the order between its
elements.
If neither a literal~$l$ nor its complement occur in $M$, then $l$ is
{\em unassigned} by $M$.


Each module $S$ determines its \emph{transition graph} ${\am}_S$:
The set of nodes of ${\am}_S$ consists of  the states relative to 
the vocabulary of~$S$. 
The edges of the
graph~${\am}_S$ are specified by the \emph{transition rules}
listed in Figure~\ref{fig:rulesam}. The first three rules depend
on the module, the fourth rule, $\rd$, does not. It has the same form 
no matter what module we consider. Hence, we omit the reference to the 
module from its notation.


\begin{figure}
$$
\begin{array}[t]{ll}

{\rp}_S:&
\quad M  ~\lrar~ 
       M~l \hbox{~ if ~$S$ has an edge from $M$ to $M~l$}\\
\phantom{.}\\
\hbox{{\rf$_S$}:}&
\quad M ~\lrar~  {\fail}
  \hbox{~ if}
  \left\{ \begin{array}{l}
  \hbox{$S$ has an edge from $M$ to $\bot$,}\\
 \hbox{$M$ contains no decision literals}
  \end{array}\right. \\
\phantom{.}\\
\hbox{{\rb$_S$}:}&
\quad P~l^\dec~Q\lrar
  P~\overline{l}
  \hbox{~ if}
  \left\{ \begin{array}{l}
\hbox{$S$ has an edge from $P~l~Q$ to $\bot$,}\\
\hbox{$Q$ contains no  decision literals}
  \end{array}\right.\\
\phantom{.}\\
\hbox{{\rd}:}&

\quad M ~\lrar~
       M~l^\dec
  \hbox{~ if  ~$l$ is unassigned by $ M$}
\end{array}
\vspace*{-0.1in}
$$

\caption{The transition rules of the graph $\am_S$.}\label{fig:rulesam}
\end{figure}

The graph ${\am}_S$ can be used to decide whether a module~$S$ has 
a model. The following properties are essential.

\begin{theorem}\label{prop:am} For
every sound module
$S$,
\begin{enumerate}
\item [(a)] graph ${\am}_S$ is finite and acyclic,
\item [(b)] for any terminal state $M$ of ${\am}_S$ other than {\fail},
$M^+$ is a model of~$\;S$,
\item [(c)] state {\fail} is reachable from $\emptyset$ in ${\am}_S$ if and
  only if $\;S$ is unsatisfiable (has no models).
\end{enumerate}
\end{theorem}

Thus, to decide whether a sound module $S$ has a model it is enough 
to find in the graph ${\am}_S$ a path leading from node $\emptyset$ 
to a terminal node $M$. If $M=\fail$, $S$ is unsatisfiable. Otherwise, 
$M$ is a model of~$S$. 

For instance, let $S$ be a module 
in  Figure~\ref{fig:modex2}. 
Below we show a 
path in the transition graph $\am_S$ with every edge annotated by 
the 
corresponding
transition rule: 
\beq
\emptyset\quad \stackrel{\rd}{\lrar}\quad b^\dec\quad
\stackrel{\rp_S}{\lrar}\quad b^\dec~\neg a\mbox{.}
\eeq{eq:patham1}
The state $b^\dec~\neg a$ is terminal. Thus, 
Theorem~\ref{prop:am} (b) asserts that 
$\{b, \neg a\}$ is a model of~$S$. There may be several paths determining the same 
model. For instance, the path
\beq
\emptyset\quad \stackrel{\rd}{\lrar}\quad \neg a^\dec\quad
\stackrel{\rd}{\lrar}\quad \neg a^\dec~b^\dec\mbox{.}
\eeq{eq:patham2}
leads to the terminal node $\neg a^\dec~b^\dec$, which is different
from $b^\dec~\neg a$ but corresponds to the same model.  
We can view a path in the graph $\am_S$  as a description of
a process of search for a model of module $S$ 
by applying transition rules. Therefore, we can characterize 
a solver based on the transition system $\am_S$
 by describing a strategy for choosing a path in 
$\am_S$. 
Such a strategy can be based, in particular, on assigning priorities to
some or all transition rules of $\am_S$, so that a solver will never
apply a transition rule in a state if a rule with higher priority is applicable
to the same state.
For example,  priorities
$$
\ba l
\rb_S,\rf_S>>
\rp_S>>
\rd\hbox{}
\ea 
$$
on the transition rules of $\am_S$
specify a solver that follows available inferences (modeled by
edges in the module $S$) before executing a transition due to {\rd}. 
The path~\eqref{eq:patham1} in the transition graph 
of the module from Figure \ref{fig:modex2} follows that strategy, 
whereas the path~\eqref{eq:patham2} does not.

We now review the graph $\dps_F$ introduced for the classical DPLL
algorithm by 
Nieuvenhuis et al.~\cite{nie06}, 
adjusting the presentation to the form 
convenient for our purposes. We then 
demonstrate its relation to the $\am_S$ graph. 
The set of nodes of ${\dps}_F$ consists of  the states relative to 
the vocabulary of a CNF formula (a set of clauses)~$F$. 
The edges of the
graph~${\dps}_F$ are specified by the transition rule $\rd$ of the
graph $\am_S$  and the rules
presented in Figure~\ref{fig:rulesdps}.
\begin{figure}
$$
\begin{array}[h]{ll}
{\rup}_F:  &
\quad M  ~\lrar~ M~l
  \hbox{~ if }
  \left\{ \begin{array}{l}
  \hbox{$C\vee l\in F$ and $M\models \neg C$,}\\ 
 \hbox{$l$ is unassigned by $M$}
  \end{array}\right. \\
\phantom{.}\\
\hbox{{\rf}$_F$:} &
\quad M ~\lrar~  {\fail}
  \hbox{~ if }
  \left\{ \begin{array}{l}
  \hbox{$C\in F$ and $M\models \neg C$,}\\
 \hbox{$M$ contains no decision literals}
  \end{array}\right. \\
\phantom{.}\\
\hbox{{\rb}$_F$:} &
\quad P~l^\dec~Q\lrar
  P~\overline{l}
  \hbox{~ if }
  \left\{ \begin{array}{l}
\hbox{$C\in F$ and $P~l^\dec~Q\models \neg C$,}\\
\hbox{$Q$ contains no  decision literals}
  \end{array}\right.\\
\end{array}
$$
\caption{Three transition rules of the graph $\dps_F$.}\label{fig:rulesdps}
\end{figure} 
For example, let $F_1$ be the theory consisting of a single clause~$a$. 
Figure~\ref{fig:modexdps2} presents $\dps_{F_1}$. 

\tikzstyle{vertex}=[draw, rectangle,  text centered,
  rounded corners]

\tikzstyle{line}=[draw, -latex']

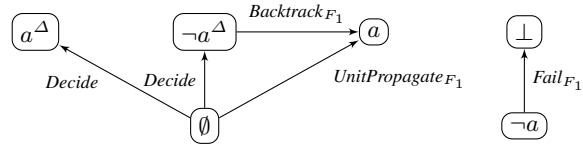
\begin{figure}[ht]
\centering{
{\footnotesize
\begin{tikzpicture}

\node [vertex] (nad) at (0,1.25) {$a^\dec$};

\node [vertex] (n0) at (2.25,0) {$\emptyset$};
\node [vertex] (nnegad) at (2.25,1.25) {$\neg a^\dec$};

\node [vertex] (na) at (4.5,1.25) {$a$};

\node [vertex] (nnega) at (6.5,0) {$\neg a$};
\node [vertex] (nbot) at (6.5,1.25) {$\bot$};

\path [line]  (n0) -- node [right]{~~~~~~{\scriptsize \rup$_{F_1}$}} (na);
\path [line] (n0) -- node[left] {{\scriptsize \rd}~~~~} (nad);
\path [line] (n0) -- node [left]{{\scriptsize \rd}} (nnegad);
\path [line] (nnega) -- node [right]{{\scriptsize \rf$_{F_1}$}} (nbot);
\path [line] (nnegad) -- node  [above] {{\scriptsize \rb$_{F_1}$}} (na);

\end{tikzpicture}
}
}
\caption{The $\dps_{F_1}$ graph where $F_1$ is a single clause $a$.}
\label{fig:modexdps2}
\end{figure}

For a CNF formula~$F$, by $\mu(\dps_F)$ we denote the graph (abstract module)
constructed from $\dps_F$ 
by  dropping all nodes that contain decision literals. 
We note that for the graph $\dps_{F_1}$ in Figure~\ref{fig:modexdps2},
the module $\mu(\dps_{F_1})$ coincides with
the module in Figure~\ref{fig:modex1}(a). This is a manifestation of
a general property.

\begin{proposition}\label{prop:dps-am}
For every CNF formula~$F$, the graph $\mu(\dps_F)$ is a sound abstract 
module equivalent to $F$. Furthermore, the graphs 
$\am_{\mu(\dps_{F})}$ 
and $\dps_{F}$ are identical.
\end{proposition}
Theorem~\ref{prop:am} and the fact that the module $\mu(\dps_F)$ is 
equivalent to a CNF formula~$F$ (Proposition \ref{prop:dps-am}) imply 
that the graph $\dps_F$ can be used for deciding the
satisfiability of~$F$. 
It is enough to find a path leading from node
$\emptyset$ to a terminal node $M$: if $M=\bot$ then~$F$ is
unsatisfiable; otherwise, $M$ is a model of~$F$. 
For instance, the only terminal states   reachable from the state
$\emptyset$ in $\dps_{F_1}$ are $a$ and $a^\dec$. This translates into the fact that $a$ is
a model of~$F_1$.
This is
exactly the result that Nieuwenhuis et al. \cite{nie06}
stated  for the graph $\dps_F$:
 \begin{corollary}\label{prop:as} For any CNF formula $F$,
\begin{enumerate}
\item [(a)] graph ${\dps}_F$ is finite and acyclic,
\item [(b)] for any terminal state $M$ of ${\dps}_F$ other than {\fail},
$M$ is a model of~$\;F$,
\item [(c)] state {\fail} is reachable from $\emptyset$ in ${\dps}_F$ if and
  only if $\;F$ is unsatisfiable (has no models).
\end{enumerate}
\end{corollary}

We now introduce the graph $\as_\Pi$ that
extends the {\dpll} graph by Nieuwenhuis et al. so that the result
can be used to specify an algorithm for finding answer sets of a
program. 
The graph $\as_\Pi$ 
can be used to form a sound module equivalent to a program~$\Pi$ in the
same way as we used $\dps_F$ to form a sound module equivalent to a CNF
formula $F$. 

We assume the reader to be familiar with the concept of \emph{unfounded 
sets}~\cite{van91,lee05}. For a set $M$ of literals and a program
$\Pi$, by $U(M,\Pi)$ we denote an unfounded set on $M$ w.r.t. $\Pi$.
It is common to identify logic rules of a program with sets of
clauses. 
By $\Pi^{cl}$ we denote the set of clauses corresponding to
the rules of $\Pi$. For instance, let $\Pi$ be  \eqref{eq:pr}, then
$\Pi^{cl}$ consists of clauses $a\vee \neg a, a\vee b$.

The set of nodes of ${\as}_\Pi$ consists of  the states relative to 
the vocabulary of program~$\Pi$. 
The edges of the
graph~${\as}_\Pi$ are specified by the transition rules of the
graph $\dps_{\Pi^{cl}}$  and the rules
presented in Figure~\ref{fig:rulessm}. 
\begin{figure}
$$
\begin{array}[t]{ll}
{\runf}_\Pi:&
\quad M  ~\lrar~ 
       M~\neg a
  \hbox{~ if }
 \left\{ \begin{array}{l}
  \hbox{$a\in U(M,\Pi)$ and}\\
 \hbox{$\neg a$ is unassigned by $M$}
  \end{array}\right. \\
\phantom{.}\\
\hbox{{\rf}$_\Pi$:}&
\quad  M ~\lrar~  {\fail}
  \hbox{~ if }
  \left\{ \begin{array}{l}
  \hbox{$a\in U(M,\Pi)$, $a\in M$, and}\\
 \hbox{$M$ contains no decision literals}
  \end{array}\right. \\
\phantom{.}\\
\hbox{{\rb}$_\Pi$:}&
\quad  P~l^\dec~Q\lrar
  P~\overline{l}
  \hbox{~ if }
  \left\{ \begin{array}{l}
  \hbox{$a\in U(P~l~Q,\Pi)$, $a\in P~l~Q$, and}\\
\hbox{$Q$ contains no  decision literals}
  \end{array}\right.\\
\end{array}
$$
\caption{Transition rules of the graph $\as_\Pi$.}\label{fig:rulessm}
\end{figure}

For a program~$\Pi$, by $\mu(\as_\Pi)$ we denote the graph
(abstract module) constructed from $\as_\Pi$ 
by removing all nodes that contain decision literals. 
\begin{proposition}\label{prop:sm-am}
For every program~$\Pi$, the graph $\mu(\as_\Pi)$ is a sound abstract
module equivalent to a program~$\Pi$ under the 
answer set
semantics. Furthermore, the graphs 
$\am_{\mu(\as_{\Pi})}$ and $\as_{\Pi}$ are identical.
\end{proposition}

From Theorem~\ref{prop:am} and the fact that $\mu(\as_\Pi)$ is an 
abstract module equivalent to an answer-set program~$\Pi$ 
it follows that the graph $\as_\Pi$ can be used for deciding 
whether $\Pi$ has an  answer set. It is enough to find a path in 
$\as_\Pi$ leading from the node $\emptyset$ to a terminal node $M$. If 
$M=\bot$ then~$\Pi$ has no
 answer sets; otherwise, $M$ is an  answer set of~$\Pi$.
 \begin{corollary}\label{prop:as2} For any program $\Pi$,
\begin{enumerate}
\item [(a)] graph ${\as}_\Pi$ is finite and acyclic,
\item [(b)] for any terminal state $M$ of ${\as}_\Pi$ other than {\fail},
$M^+$ is an answer set of $\Pi$,
\item [(c)] state {\fail} is reachable from $\emptyset$ in ${\as}_\Pi$ if and
  only if $\Pi$ has no answer sets.
\end{enumerate}
\end{corollary}

Let $\Pi$
be the program~\eqref{eq:pr}. Figure~\ref{fig:modex3}(b)
presents the module $\mu({\as_{\Pi}})$. 
It is easy to see that 
this module is equivalently contained in the 
saturated module for $\Pi$ 
presented
 in Figure~\ref{fig:modex2}. For program $\Pi$ 
the inference rules of
 {\rup} and {\runf} are capable to capture all but one inference due to
the entailment (the missing inference corresponds to the edge from $b$ 
to $\neg a~b$ in Figure~\ref{fig:modex2}).

Let us now consider the graph $\as^-_\Pi$ constructed from $\as_\Pi$ by
either dropping the rules  $\runf_\Pi$, $\rb_\Pi$, $\rf_\Pi$ or the
rules  $\rup_{\Pi^{cl}}$, $\rb_{\Pi^{cl}}$, $\rf_{\Pi^{cl}}$. 
In each case, the module $\mu(\as^-_\Pi)$ in general is not equivalent to 
a program~$\Pi$. This demonstrates the importance of two kinds of 
inferences for the case of logic programs: (i) those stemming from 
unit propagate and related to the fact that an answer set of a program is 
also its classical model; as well as (ii) those based on the concept of 
``unfoundedness'' and related 
to the fact that every answer set of a program contains no unfounded sets. 
We note that forward chaining mentioned in earlier section is subsumed by 
unit propagate.

The graph $\as_\Pi$ is inspired by the graph  $\sm_\Pi$ introduced by
Lierler~\cite{lier08}   for specifying
an answer set solver {\sc smodels}~\cite{nie00}. The graph $\sm_\Pi$ extends 
 $\as_\Pi$ by two additional transition rules 
(inference rules or propagators): {\rarc} and {\rbt}. 
We chose to start the presentation with the graph $\as_\Pi$ for its
simplicity. We now recall the definition of $\sm_\Pi$ and illustrate
how a similar result to Proposition~\ref{prop:sm-am} is applicable to
it.

If $B$ is a conjunction of literals then by $\ol{B}$ we understand the
set of the complements of literals occurring in $B$.

The set of nodes of ${\sm}_\Pi$ consists of  the states relative to 
the vocabulary of program~$\Pi$. 
The edges of the
graph~${\sm}_\Pi$ are specified by the transition rules of the
graph $\as_{\Pi}$  and the following rules:
{\small
$$
\begin{array}[t]{ll}
{\rarc}:&
\quad M  ~\lrar~ 
       M~\neg a
  \hbox{~ if }
  \left\{ \begin{array}{l}
  \hbox{$\ol{B}\cap M\neq\emptyset$ for all $B\in Bodies(\Pi,a)$}\\
 \hbox{$\neg a$ is unassigned by $M$}
  \end{array}\right. \\
\phantom{.}\\
\hbox{{\rfarc}$$:}&
\quad  M ~\lrar~  {\fail}
  \hbox{~ if }
  \left\{ \begin{array}{l}
  \hbox{$\ol{B}\cap M\neq\emptyset$ for all $B\in Bodies(\Pi,a)$,}\\
 \hbox{$a\in M$, $M$ contains no decision literals}

  \end{array}\right. \\
\phantom{.}\\
\hbox{{\rbarc}$$:}&
\quad  P~l^\dec~Q\lrar
  P~\overline{l}
  \hbox{~ if }
  \left\{ \begin{array}{l}
  \hbox{$\ol{B}\cap M\neq\emptyset$ for all $B\in Bodies(\Pi,a)$, }\\
 \hbox{$a\in P~l~Q$, $Q$ contains no  decision literals}
  \end{array}\right.\\
\phantom{.}\\
{\rbt}:&
\quad M  ~\lrar~ 
       M~l
  \hbox{~ if }
  \left\{ \begin{array}{l}
  \hbox{$a\ar B\in \Pi$, $a\in M$, $l\in B$}\\
 \hbox{$\ol{B'}\cap M\neq\emptyset$ for all $B'\in Bodies(\Pi,a)\setminus B$}\\
 \hbox{$l$ is unassigned by $M$}
  \end{array}\right. \\
\phantom{.}\\
\hbox{{\rfbt}$$:}&
\quad  M ~\lrar~  {\fail}
  \hbox{~ if }
  \left\{ \begin{array}{l}
  \hbox{$a\ar B\in \Pi$, $a\in M$, $l\in B$}\\
 \hbox{$\ol{B'}\cap M\neq\emptyset$ for all $B'\in Bodies(\Pi,a)\setminus B$}\\
 \hbox{$l\in M$, $M$ contains no decision literals}\\
  \end{array}\right. \\
\phantom{.}\\
\hbox{{\rbbt}$$:}&
\quad  P~l^\dec~Q\lrar
  P~\overline{l}
  \hbox{~ if }
  \left\{ \begin{array}{l}
  \hbox{$a\ar B\in \Pi$, $a\in P~l~Q$, $l'\in B$}\\
 \hbox{$\ol{B'}\cap P~l~Q\neq\emptyset$ for all $B'\in Bodies(\Pi,a)\setminus B$}\\
 \hbox{$l'\in P~l~Q$, $Q$ contains no decision literals}\\
  \end{array}\right.\\
\end{array}
$$
}

The graph $\sm_\Pi$ shares the important properties of the graph $\as_\Pi$.
Indeed, Proposition~\ref{prop:sm-am} and Corollary~\ref{prop:as2} hold
if one replaces $\as_\Pi$ with $\sm_\Pi$. Corollary~\ref{prop:as2} in this
form was one
of the main results stated in~\cite{lier08}\footnote{In~\cite{lier08}, Lierler
presented the $\sm_\Pi$ graph in a slightly different from: the states of that
graph permitted inconsistent states of literals, which in turn allowed
to unify the {\rf} and {\rb} transition rules for different propagators.}.

Let $\Pi$
be the program~\eqref{eq:pr}. Figure~\ref{fig:modex3}(b)
presents the module $\mu({\as_{\Pi}})$. 
The module $\mu({\sm_{\Pi}})$ coincides with
the 
saturated module for $\Pi$ 
presented
 in Figure~\ref{fig:modex2}.
 For program~$\Pi$, 
the inference rule 
 {\rbt} captures the inference 
that corresponds to the edge from $b$ 
to $\neg a~b$, which the transition rules of the graph ${\as_{\Pi}}$
are incapable to capture.

The examples above show that the framework of abstract modules uniformly 
encompasses different logics. We illustrated this point by means of propositional logic and answer-set 
programming. Furthermore, it uniformly  models diverse reasoning mechanisms
(entailment and its proof theoretic specializations). The results also
demonstrate that transition systems proposed earlier to represent and analyze 
SAT and ASP solvers are special cases of general transition systems for abstract modules
introduced here. 

\section{Abstract Modular System and Solver $\ams_\cS$}\label{sec:ams}

By capturing diverse logics in a single framework, abstract modules are 
well suited for studying modularity in declarative formalisms, and
principles underlying solvers for modular declarative formalisms.
We now define an abstract modular declarative framework that uses the 
concept of a module as its basic element. We then show how abstract
transition systems for modules generalize to the new formalism.

An \emph{abstract modular system} (AMS) is a set of modules. The 
vocabulary of an AMS~$\cS$ is the union of the vocabularies of
modules of $\cS$ (they do not have to have the same vocabulary);
we denote it
by $\sigma(\cS)$. 

An interpretation $I$ over $\sigma(\cS)$ (that is, a subset of $\sigma(\cS)$)
is a \emph{model} of~$\cS$, written $I\models\cS$,
if $I$ is a model of every module~$S\in \cS$.
An AMS $\cS$ \emph{entails} a formula $\vph$ (over the same
vocabulary as $\cS$), written $\cS\models\vph$, if for every model
$I$ of $\cS$ we have $I\models \vph$. 
We say that an AMS $\cS$ is {\em sound} if every module~$S\in \cS$ is sound.

Let $S_1$ be a module presented in
Figure~\ref{fig:modex1}(a) and $S_2$ be a module in
Figure~\ref{fig:modex3}(b). The vocabulary of the AMS $\{S_1,S_2\}$
consists of the atoms $a$ and $b$. It is easy to see that the interpretation $\{a, \neg b\}$
is its only model. 

For a vocabulary~$\sigma$ and a set of literals $M$, by $M|_\sigma$ we
denote the maximal subset of~$M$ consisting of literals over~$\sigma$. For
example,  $\{\neg a,\neg b\}|_{\{a\}}=\{\neg a\}$.

Each AMS $\cS$ determines its \emph{transition system} $\ams_\cS$.
The set of nodes of $\ams_\cS$ consists of the states relative to $\sigma(\cS)$.
The transition rules of $\ams_\cS$ comprise the rule $\rd$ and 
the rules ${\rp}_S$, $\rf_S$, and $\rb_S$, for all modules $S \in
\cS$. The latter three 
rules are modified to account for the vocabulary $\sigma(\cS)$ and 
 are presented in Figure~\ref{fig:rulesamc}.

\begin{figure}
$$
\begin{array}[t]{ll}
{\rp}_S:  &
\quad M  ~\lrar~ 
       M~l \hbox{~ if ~~$S$ has an edge from $M|_{\sigma(S)}$ to $M~l|_{\sigma(S)}$}\\
\phantom{.}\\
\hbox{{$\rf_S$}:} &
\quad  M ~\lrar~  {\fail}
  \hbox{~ if }
  \left\{ \begin{array}{l}
  \hbox{$S$ has an edge from $M|_{\sigma(S)}$ to $\bot$,}\\
 \hbox{$M$ contains no decision literals}
  \end{array}\right. \\
\phantom{.}\\
\hbox{{\rb$_S$}:} &
\quad  P~l^\dec~Q\lrar
  P~\overline{l}
  \hbox{~ if }
  \left\{ \begin{array}{l}
\hbox{$S$ has an edge from $P~l~Q|_{\sigma(S)}$ to $\bot$,}\\
\hbox{$Q$ contains no  decision literals}
  \end{array}\right.\\
\phantom{.}
\end{array}
\vspace*{-0.1in}
$$
\caption{The transition rules of the graph $\ams_\cS$.}\label{fig:rulesamc}
\end{figure}

\begin{theorem}\label{prop:ams} For
  every sound AMS  $\cS$,
\begin{itemize}
\item [(a)] the graph ${\ams}_\cS$ is finite and acyclic,
\item [(b)] any terminal state of ${\ams}_\cS$ other than {\fail} 
is a model of~$\cS$,
\item [(c)] the state {\fail} is reachable from $\emptyset$ in ${\ams}_\cS$ if and
  only if $\cS$ is unsatisfiable.
\end{itemize}
\end{theorem}
This theorem demonstrates that to decide a satisfiability of a
sound AMS $\cS$ it is
sufficient to find a path leading from node $\emptyset$ to a terminal node.
It provides a foundation for the development and analysis
of solvers for modular systems. 

For instance, let $\cS$ be the AMS $\{S_1,S_2\}$. Below is a valid
path in the transition graph $\ams_\cS$ with every edge annotated 
by the 
corresponding
transition rule: 
$$
\emptyset \stackrel{\rd}{\lrar} \neg a^\dec
\stackrel{\rp_{S_2}}{\lrar}
\neg~a^\dec~b\stackrel{\rb_{S_1}}{\lrar}  a
\stackrel{\rd}{\lrar} a~\neg b^\dec
\mbox{.}
$$
The state $a~\neg b^\dec$ is terminal. Thus, 
Theorem~\ref{prop:ams} (b) asserts that 
$\{a,\neg b\}$ is a model of~$\cS$. 
Let us interpret this example. Earlier we demonstrated that module 
$S_1$ can be regarded as a representation of a propositional theory 
consisting of a single clause $a$ whereas $S_2$ corresponds to the
logic program~\eqref{eq:pr} under the semantics of  answer sets. 
We then illustrated how modules $S_1$ and $S_2$ give rise to particular
algorithms for implementing search procedures. The graph $\ams_\cS$ 
represents the algorithm obtained by \emph{integrating} the algorithms 
supported by the modules $S_1$ and $S_2$ separately.

The results presented above imply, as special cases, 
earlier results on the logics PC(ID) and SM(ASP), and their solvers
\cite{lt2011}.



\section{Learning in Solvers for AMSs.}
Nieuwenhuis et al.~\cite[Section~2.4]{nie06}  defined 
 the \emph{DPLL System with Learning} graph 
to describe 
SAT solvers' 
learning, 
one of the crucial features of
current SAT solvers responsible for rapid success in this area
of automated reasoning. 
The approach of Nieuwenhuis, Oliveras, and Tinelli extends to our 
abstract setting. 
Specifically, the graph $\ams_\cS$ can be extended with ``learning 
transitions'' to represent solvers for AMSs that 
incorporate learning.

The intuition behind learning in SAT is to allow new propagations 
by extending the original clause database as computation proceeds.
These ``learned'' clauses provide new ``immediate derivations'' 
to a SAT solver by enabling additional applications of {\rup}. 
In the framework of abstract modules, immediate derivations are 
represented by edges. Adding edges to modules captures the idea of learning
by supporting new propagations that the transition rule {\rp} may take 
an advantage of. We now state these intuitions formally for the case 
of abstract modular systems. 

Let $S$ be a module and $E$ a set of edges between nodes of $S$.
By $S^E$ we 
denote the module constructed by adding to $S$ the edges in $E$.
A set $E$ of edges is \emph{$S$-safe}
if the module $S^E$ is sound and equivalent to $S$.
For an AMS $\cS$ and a set of edges $E$ over the vocabulary of $\cS$, 
we define $\cS^E = \{S^{E_{|S}}\colon S\in \cS\}$ (where $E_{|S}$ is the set
of those edges in $E$ that connect nodes in $S$). We say that $E$ is 
\emph{$\cS$-safe} if $\cS$ and $\cS^E$ are \emph{equivalent}, and each 
module $S^E$ in $\cS^E$ is sound.

An {\em (augmented) state} relative to an AMS $\cS=\{S_1, \dots, S_n\}$ is 
either a distinguished state {\fail} or a pair of the form 
$M||\Gamma_1,\dots,\Gamma_n$ where $M$ is an \emph{ordered} consistent set~$M$ of literals over $\sigma$, some possibly
annotated by $\dec$; and $\Gamma_1, \dots,\Gamma_n$ are sets of edges 
between nodes of modules $S_1,\ldots,S_n$, respectively. 
Sometimes we denote $\Gamma_1,\dots,\Gamma_n$ by $\mg$. 
For any AMS $\cS=\{S_1,\dots,S_n\}$, we define a graph $\amsl_\cS$. Its 
nodes are the augmented states relative to~$\cS$. The  rule 
{\rd} of the $\ams_\cS$ graph extends to $\amsl_\cS$ as follows 
{\small
$$
\ba{ll}
\hbox{{\rd}:}&
\quad M||\mg ~\lrar~
       M~l^\dec||\mg
  \hbox{~ if  ~~$l$ is unassigned by $M$.}
\ea
$$
}
Figure~\ref{fig:rulesamsl} presents the transition rules of
$\amsl_\cS$ that are specific to each module $S_i$ in~$\cS$.
We note that the set $E$ of edges in the rule $\rll_{S_i}$ is required to 
consist of edges that run between the nodes of $S_i$.
The transition rule 
{\small
$$
\ba {ll}
\hbox{{\rlg}:}&
\quad  M ||\dots,\Gamma_j,\dots~\lrar~ 
       M||\dots,\Gamma_j\cup E_{|S_i},\dots ~
  \hbox{if $E$ is $\cS$-safe}  
\ea
$$
}
where $E$ is a set of edges between nodes over the vocabulary $\sigma(\cS)$,
concludes the definition of $\amsl_\cS$.

\begin{figure}
$$
\begin{array}[t]{ll}

{\rp}_{S_i}: &
\quad  M||\mg  ~\lrar~ 
       M~l||\mg \hbox{~if ~$S_i^{\Gamma_i}$ has an edge from $M$ to $M~l$}\\
\phantom{.}\\
\hbox{{\rf$_{S_i}$}:}&
\quad  M||\mg ~\lrar~  {\fail}
  \hbox{~if}
  \left\{ \begin{array}{l}
  \hbox{$S_i^{\Gamma_i}$ has an edge from $M$ to $\bot$,}\\
 \hbox{$M$ contains no decision literals}
  \end{array}\right. \\
\phantom{.}\\
\hbox{{\rb$_{S_i}$}:} &
\quad  P~l^\dec~Q||\mg\lrar
  P~\overline{l}||\mg
  \hbox{~if}
  \left\{ \begin{array}{l}
\hbox{$S_i^{\Gamma_i}$ has an edge from $P~l~Q$ to $\bot$,}\\
\hbox{$Q$ contains no  decision literals}
  \end{array}\right.\\
\phantom{.}\\
\hbox{{\rll$_{S_i}$}:}&
\quad  M ||\dots,\Gamma_i,\dots~\lrar~ 
       M||\dots,\Gamma_i\cup E,\dots ~
  \hbox{ if $E$ is $S_i$-safe~}
\end{array}
\vspace*{-0.1in}
$$
\caption{Transition rules of $\amsl_\cS$ for module
  $S_i\in S$.}\label{fig:rulesamsl}
\end{figure}

We refer to the transition rules \rp,
\rb, {\rd}, and {\rf} of the graph  $\amsl_\cS$  as {\em basic}.
We say that a node in the graph is {\em semi-terminal} 
if no basic rule  is applicable to it.
The graph $\amsl_\cS$ can be used  for deciding
whether an AMS~$\cS$ has an answer set by constructing a path from
$\emptyset||\emptyset,\dots,\emptyset$ to a semi-terminal node.

\begin{theorem}\label{prop:amsl} For any sound AMS $\cS$,
\begin{itemize}
\item [(a)] 
there is an integer $m$ such that every path in ${\amsl}_\cS$ contains 
at most $m$ edges due to basic transition rules,
\item [(b)] for any  semi-terminal state $M||\mg$  of
  ${\amsl}_\cS$ reachable from $\emptyset||\emptyset,\dots,\emptyset$, 
$M$ is a model of~$\;\cS$,
\item [(c)] state {\fail} is reachable from
  $\emptyset||\emptyset,\dots,\emptyset$ 
in ${\amsl}_\cS$ if and
  only if $\;\cS$ has no models.
\end{itemize}
\end{theorem}

It follows that if we are constructing a path starting in 
$\emptyset||\emptyset,\dots, \emptyset$ in a way that guarantees that 
every sequence of consecutive edges of the path labeled with $\rll$ 
and $\rlg$ eventually ends (is finite), then the path will reach some 
semi-terminal state. As soon as a semi-terminal state is reached the 
problem of finding a model is solved.

There is an important difference between {\rll} and {\rlg}. The 
first one allows new propagations within a module but does not change 
its semantics as the models of the module stay the same (and it is 
local, other modules are unaffected by it). The application of {\rlg} 
while preserving the overall semantics of the system may change the 
semantics of individual modules by eliminating some of their models 
(and, being global, affects in principle all modules of the system). 
SAT researchers have demonstrated that {\rll} is crucial for the success
of SAT technology both in practice and theoretically. Our initial
considerations suggest that under some circumstances, {\rlg} offers 
additional substantial performance benefits.

We stress that our discussion of learning does not aim at any specific
algorithmic ways in which one could perform learning. Instead, we 
formulate conditions that learned edges are to satisfy ($S$-safety for 
learning local to a module $S$, and $\mathcal{A}$-safety for the global learning 
rule), which ensure the correctness of solvers that implement learning 
so that to satisfy them. In this way, we provide a uniform framework
for correctness proofs of multi-logic solvers incorporating learning.

\section{Related Work}\label{sec:related}
In an important development, Brewka and Eiter~\cite{bre07} introduced 
an abstract notion of a \emph{heterogeneous nonmonotonic multi-context 
system} (MCS).
One of the key aspects of that 
proposal is its abstract representation of a logic and hence contexts
that rely on such abstract logics. The independence of contexts from
syntax promoted focus on semantic aspect of modularity exhibited by
multi-context systems.
Since their inception,
multi-context systems have received substantial attention and inspired
implementations of hybrid reasoning systems including {\sc
  dlvhex}~\cite{eit05a} and {\sc dmcs}~\cite{bai10}. 
Abstract modular systems introduced here are similar to MCSs 
as they too do not rely on any particular  syntax for
logics assumed in modules (a counterpart of a context). What distinguishes 
them is that they
encapsulate some semantic features
stemming from inferences allowed by the underlying logic. This feature of abstract
modules is essential for our purposes as we utilize them as a tool for studying
algorithmic aspects of multi-logic systems. Another difference
between AMS and MCS is due to  ``bridge rules.'' Bridge rules  are 
crucial for defining the semantics of an MCS. They are also responsible
for ``information sharing'' in MCSs. They are absent in our
formalism altogether. In AMS information sharing is implemented by a simple
notion of a shared vocabulary between the modules.

Modularity is one of the key techniques in principled software 
development. This has been a major trigger inspiring research on
modularity in declarative programming
paradigms rooting in KR languages such as answer set programming, for instance.
Oikarinen and Janhunen~\cite{oik06}  proposed a modular version 
of answer set programs called lp-modules. In that work, the authors were primarily concerned
with the decomposition of lp-modules into sets of simpler ones. They
proved that under some assumptions such decompositions are possible.
J\"{a}rvisalo, Oikarinen, Janhunen, and Niemel\"a~\cite{oik09}, and Tasharrofi and Ternovska~\cite{tas11} studied the generalizations of lp-modules. In 
their work the main focus was to abstract lp-modules formalism away from
any particular syntax or semantics. They then study properties of the
modules such as ``joinability'' and analyze
{\em different ways} to join modules together and the semantics of
such a join.
We are interested in 
building simple modular systems using abstract modules -- the only
composition mechanism that we study is based on conjunction of
modules. Also in contrast to the work by J\"{a}rvisalo et al.~\cite{oik09} 
and Tasharrofi and Ternovska~\cite{tas11}, 
we define such conjunction for any modules disregarding
their internal structure and interdepencies between each other. 

Tasharrofi, Wu, and Ternovska~\cite{tas11b} developed and studied an 
algorithm for processing modular model expansion tasks in the abstract
multi-logic system concept developed by Tasharrofi and Ternovska~\cite{tas11}.
They use the traditional pseudocode method to present the developed 
algorithm. In this work we adapt the graph-based framework for designing 
backtrack search algorithms for abstract modular systems. The benefits of
that approach for modeling families of backtrack search procedures employed 
in SAT, ASP, and PC(ID) solvers were demonstrated by 
Nieuwenhuis et al.~\cite{nie06}, 
Lierler~\cite{lier08}, and
Lierler and Truszczynski~\cite{lt2011}. 
Our work provides additional support for the generality and flexibility
of the graph-based framework as a finer abstraction of backtrack search
algorithms than direct pseudocode representations, allowing for convenient
means to prove correctness and study relationships between the families 
of the algorithms.


\section{Conclusions}

We introduced abstract modules and abstract modular systems and showed
that they provide a framework capable of capturing diverse logics and 
inference mechanisms integrated into modular knowledge representation 
systems. In particular, we showed that transition graphs determined by 
modules and modular systems provide a unifying representation of 
model-generating algorithms, or solvers, and simplify reasoning about 
such issues as correctness or termination. We believe they can be 
useful in theoretical comparisons of solver effectiveness and in
the development of new solvers. Learning, a fundamental technique in 
solver design, displays itself in two quite different flavors, local and 
global. The former corresponds to learning studied before in SAT and SMT
and demonstrated both theoretically and practically to be essential for
good performance. Global learning is a new concept that we identified in
the context of modular systems. It concerns learning \emph{across} modules
and, as local learning, promises to lead to performance gains. In the 
future work we will conduct a systematic study of global learning in 
abstract modular systems and its impact on solvers for practical 
multi-logic formalisms. 

\bibliographystyle{splncs03}

\bibliography{abstractmods-bib}

\begin{thebibliography}{10}
\providecommand{\url}[1]{\texttt{#1}}
\providecommand{\urlprefix}{URL }

\bibitem{bai10}
Bairakdar, S.E.D., Dao-Tran, M., Eiter, T., Fink, M., Krennwallner, T.: The
  dmcs solver for distributed nonmonotonic multi-context systems. In: 12th
  European Conference on Logics in Artificial Intelligence (JELIA). pp.
  352--355 (2010)

\bibitem{BarretSST08}
Barrett, C., Sebastiani, R., Seshia, S., Tinelli, C.: Satisfiability modulo
  theories. In: Biere, A., Heule, M., van Maaren, H., Walsch, T. (eds.)
  Handbook of Satisfiability, pp. 737--797. IOS Press (2008)

\bibitem{bre07}
Brewka, G., Eiter, T.: Equilibria in heterogeneous nonmonotonic multi-context
  systems. In: Proceedings of National conference on Artificial Intelligence
  ({AAAI}). pp. 385--390 (2007)

\bibitem{dltj12}
Denecker, M., Lierler, Y., Truszczynski, M., Vennekens, J.: A {T}arskian
  informal semantics for answer set programming. In: Dovier, A., Costa, V.S.
  (eds.) International Conference on Logic Programming (ICLP). LIPIcs, vol.~17.
  Schloss Dagstuhl - Leibniz-Zentrum fuer Informatik (2012)

\bibitem{eit05a}
Eiter, T., Ianni, G., Schindlauer, R., Tompits, H.: A uniform integration of
  higher-order reasoning and external evaluations in answer set programming.
  In: Proceedings of International Joint Conference on Artificial Intelligence
  ({IJCAI}). pp. 90--96 (2005)

\bibitem{gel88}
Gelfond, M., Lifschitz, V.: The stable model semantics for logic programming.
  In: Kowalski, R., Bowen, K. (eds.) Proceedings of International Logic
  Programming Conference and Symposium. pp. 1070--1080. MIT Press (1988)

\bibitem{Giunchiglia93}
Giunchiglia, F.: Contextual reasoning. Epistemologia  XVI,  345--364 (1993)

\bibitem{jm94}
Jaffar, J., Maher, M.: Constraint logic programming: A survey. Journal of Logic
  Programming  19(20),  503--581 (1994)

\bibitem{oik09}
J\"{a}rvisalo, M., Oikarinen, E., Janhunen, T., Niemel\"{a}, I.: A module-based
  framework for multi-language constraint modeling. In: Proceedings of the 10th
  International Conference on Logic Programming and Nonmonotonic Reasoning. pp.
  155--168. LPNMR '09, Springer-Verlag, Berlin, Heidelberg (2009),
  \url{http://dx.doi.org/10.1007/978-3-642-04238-6_15}

\bibitem{lee05}
Lee, J.: A model-theoretic counterpart of loop formulas. In: Proceedings of
  International Joint Conference on Artificial Intelligence ({IJCAI}). pp.
  503--508. Professional Book Center (2005)

\bibitem{lier08}
Lierler, Y.: Abstract answer set solvers. In: Proceedings of International
  Conference on Logic Programming (ICLP). pp. 377--391. Springer (2008)

\bibitem{lier10}
Lierler, Y.: Abstract answer set solvers with backjumping and learning. Theory
  and Practice of Logic Programming  11,  135--169 (2011)

\bibitem{lier12aaai}
Lierler, Y.: On the relation of constraint answer set programming languages and
  algorithms. In: Proceedings of the AAAI Conference on Artificial
  Intelligence. MIT Press (2012)

\bibitem{lt2011}
Lierler, Y., Truszczynski, M.: Transition systems for model generators --- a
  unifying approach. Theory and Practice of Logic Programming, 27th Int'l.
  Conference on Logic Programming (ICLP'11) Special Issue  11, issue 4-5 (2011)

\bibitem{mar99}
Marek, V., Truszczy\'nski, M.: Stable models and an alternative logic
  programming paradigm. In: The Logic Programming Paradigm: a 25-Year
  Perspective, pp. 375--398. Springer Verlag (1999)

\bibitem{mar08}
Mari{\"e}n, M., Wittocx, J., Denecker, M., Bruynooghe, M.: {S}{A}{T}({I}{D}):
  Satisfiability of propositional logic extended with inductive definitions.
  In: SAT. pp. 211--224 (2008)

\bibitem{mcc87}
McCarthy, J.: Generality in {A}rtificial {I}ntelligence. Communications~of the
  ACM  30(12),  1030--1035 (1987), reproduced in \cite{mcc90}

\bibitem{nie99}
Niemel{\"a}, I.: Logic programs with stable model semantics as a constraint
  programming paradigm. Annals of Mathematics and Artificial Intelligence  25,
  241--273 (1999)

\bibitem{nie00}
Niemel{\"a}, I., Simons, P.: Extending the {Smodels} system with cardinality
  and weight constraints. In: Minker, J. (ed.) Logic-Based Artificial
  Intelligence, pp. 491--521. Kluwer (2000)

\bibitem{nie06}
Nieuwenhuis, R., Oliveras, A., Tinelli, C.: Solving {S}{A}{T} and {S}{A}{T}
  modulo theories: From an abstract {D}avis-{P}utnam-{L}ogemann-{L}oveland
  procedure to {D}{P}{L}{L}({T}). Journal of the ACM  53(6),  937--977 (2006)

\bibitem{oik06}
Oikarinen, E., Janhunen, T.: Modular equivalence for normal logic programs. In:
  17th European Conference on Artificial Intelligence(ECAI). pp. 412--416
  (2006)

\bibitem{ros08}
Rossi, F., van Beek, P., Walsh, T.: Constraint programming. In: van Harmelen,
  F., Lifschitz, V., Porter, B. (eds.) Handbook of Knowledge Representation,
  pp. 181--212. Elsevier (2008)

\bibitem{sns02}
Simons, P., Niemel{\"a}, I., Soininen, T.: Extending and implementing the
  stable model semantics. Artificial Intelligence  138,  181--234 (2002)

\bibitem{tas11}
Tasharrofi, S., Ternovska, E.: A semantic account for modularity in
  multi-language modelling of search problems. In: Frontiers of Combining
  Systems, 8th International Symposium (FroCoS). pp. 259--274 (2011)

\bibitem{tas11b}
Tasharrofi, S., Wu, X.N., Ternovska, E.: Solving modular model expansion tasks.
  CoRR  abs/1109.0583 (2011)

\bibitem{van91}
Van~Gelder, A., Ross, K., Schlipf, J.: The well-founded semantics for general
  logic programs. Journal of ACM  38(3),  620--650 (1991)

\end{thebibliography}

\end{document}